\documentclass[twoside,11pt]{article}

\usepackage{jmlr2e}

\usepackage{mathrsfs}
\usepackage{jmlr2e}
\usepackage{amssymb, amsmath, comment, xspace, enumerate}
\usepackage{graphics, graphicx, subfigure}
\usepackage{microtype,color}
\usepackage[normalem]{ulem}
\usepackage[makeroom]{cancel}
\usepackage{parskip}
\usepackage{mathrsfs}
\usepackage{latexsym}
\usepackage{csquotes}
\usepackage{booktabs} 

\def\<{\begin{equation}}
\def\>{\end{equation}}

\usepackage[mathcal]{eucal}

\newtheorem{thm2}{Theorem}

\newtheorem{mainidea}[thm2]{Main Idea}

\ShortHeadings{Unnormalized Variational Bayes}{Saremi}
\firstpageno{1}

\begin{document}

\title{Unnormalized Variational Bayes}

\author{\name Saeed Saremi \email saeed@berkeley.edu \\
       \addr 
NNAISENSE, Inc.\\
1224 East 12th St., suite 313 \\
Austin, TX 78702, USA \\
University of California, Berkeley
}
\editor{}

\maketitle
\begin{abstract}
We unify empirical Bayes and variational Bayes for approximating unnormalized densities. This framework, named unnormalized variational Bayes (UVB), is based on formulating a latent variable model for the random variable $Y=X+N(0,\sigma^2 I_d)$ and using the evidence lower bound (ELBO), computed by a variational autoencoder, as a parametrization of the energy function of $Y$ which is then used to estimate $X$ with the empirical Bayes least-squares estimator. In this intriguing setup, the \emph{gradient} of the ELBO with respect to noisy inputs plays the central role in learning the energy function. Empirically, we demonstrate that UVB has a higher capacity to approximate energy functions than the parametrization with MLPs as done in neural empirical Bayes (DEEN). We especially showcase $\sigma=1$, where the differences between UVB and DEEN become visible and qualitative in the denoising experiments. For this high level of noise, the distribution of $Y$ is very smoothed and we demonstrate that one can traverse in a single run\textemdash without a restart\textemdash all MNIST classes in a variety of styles via walk-jump sampling with a fast-mixing Langevin MCMC sampler. We finish by probing the encoder/decoder of the trained models and confirm UVB $\neq$ VAE.
\end{abstract}







\section{Introduction}
At a high level, this work brings together \emph{empirical Bayes} as formulated by~\citet{robbins1956empirical} together with its recent interpretation for learning unnormalized densities~\citep{saremi2019neural} and \emph{variational Bayes} in its modern formulation~\citep{jordan1999introduction, kingma2013auto, rezende2014stochastic}. This unification as it will become clear is quite general but it has been particularly inspired by the problem of learning the energy function of the random variable $Y=X+N(0,\sigma^2 I_d)$ based on the i.i.d. sequence $x_1,\dots,x_n$ sampled from $X$ in $\mathbb{R}^d$. Our plan for the introduction is to first briefly review empirical Bayes and the empirical Bayes denoising methodology for learning unnormalized densities. We then present our main contribution in bringing variational Bayes and variational autoencoders to the picture, together with a highlight of the main empirical results in the paper.

\subsection{Empirical Bayes} The story goes back to~\citep{robbins1956empirical}, where one starts with the i.i.d. sequence $y_1,\dots,y_n$ (the noisy observations that one measures) and the simple yet profound question is whether one can estimate $X$ given a single noisy measurement $Y=y$ using the least-squares estimator of $X$ assuming that \emph{we do not know the distribution of $X$}, i.e. we only know the measurement kernel $p(y|x)$ also known as the noise model. There are two main results. The first one is not surprising and is the fact that the least-squares estimator of $X$ is the Bayes estimator:
\< \label{eq:robbins} \widehat{x}(y)= \frac{\int x p(y|x) p(x) dx}{\int p(y|x) p(x) dx}  \>
The second result is remarkable and it is the fact that the Bayes estimator can be written in closed form purely based on the distribution of $Y$. For Gaussian kernels the estimator takes the form~\citep{miyasawa1961empirical}:
\< \label{eq:miyasawa} \widehat{x}(y) = y + \sigma^2 \nabla \log p(y).\>
This result is indeed surprising since, on its face, one would expect that knowing $p(x)$ is a \emph{must} to carry out the integral in~(\ref{eq:robbins}) in closed form but that is in fact not necessary for a large class of noise models; see~\citep{raphan2011least} for literature review and a unified formalism on empirical Bayes estimators. 
\begin{remark} [on notation]
	A derivation of Miyasawa's result (Eq.~\ref{eq:miyasawa}) consistent with the notations here is given in~\citep{saremi2019neural}. The only difference is that in this paper we switch from denoting densities by $``f"$ (used in the mathematical statistics literature) to $``p"$ (common in the variational Bayes literature). In this work, we are moving towards variational Bayes and this switch from $f$ to $p$ is appropriate. On the other hand, we use $f$ to denote energy functions.  In addition, we follow the convention of dropping the subscripts from densities when the arguments to the density functions are present: $p(y)=p_Y(y)$, etc.
\end{remark} 

\subsection{Neural Empirical Bayes}

With this background on empirical Bayes, consider a different problem: approximating $\nabla \log p(y)$ based on the i.i.d. sequence $x_1,\dots,x_n$. This is a simpler problem than~\citep{robbins1956empirical} since we now observe the clean samples. It is also a relaxation of the problem of estimating the density of $X$ since, \emph{to say the least},  $p_Y=p_X * \mathcal{N}$ is smoother than $p_X$. In addition, in approximating the score function $\nabla \log p(y)$ we only require approximating $\log p(y)$ modulo an additive constant. In other words, the learned density is unnormalized.

\begin{figure}[h!] 
\begin{center}
\hspace{1.5cm}
\begin{subfigure}[$Y=X+N(0,\sigma^2 I_d)$]
 {\includegraphics[width=0.29\textwidth]{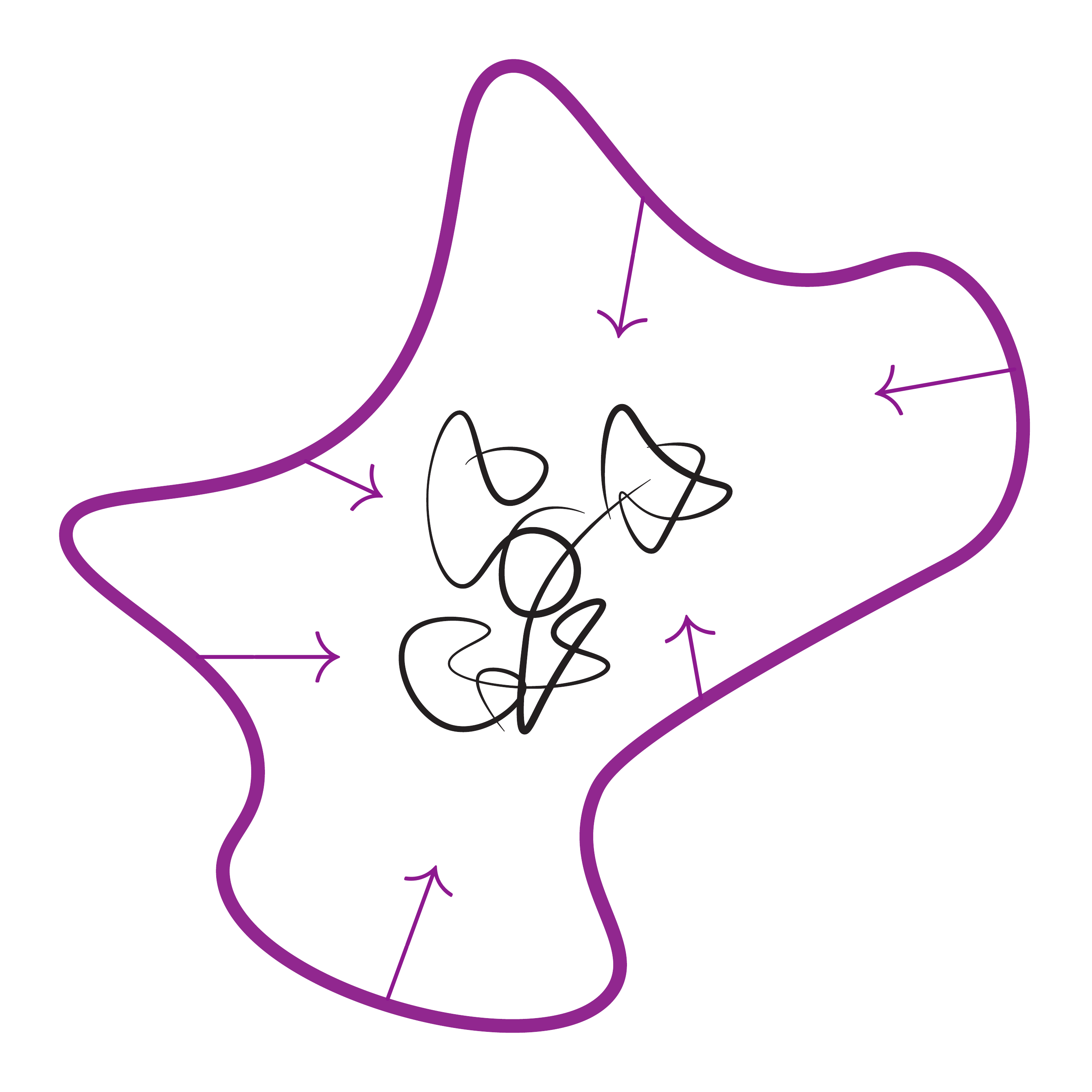}}
\end{subfigure}
\begin{subfigure}[$Z$]
 {\includegraphics[width=0.29\textwidth]{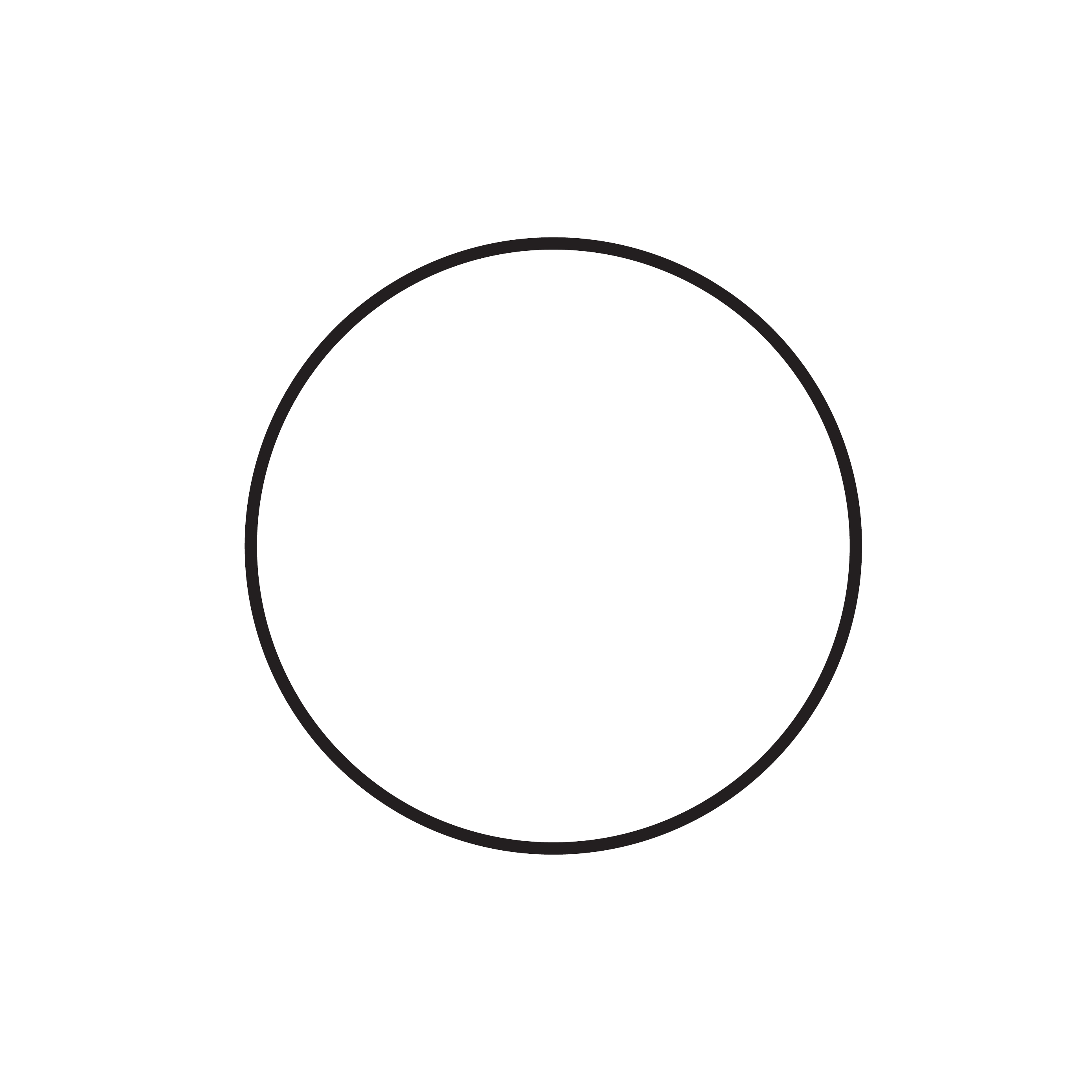}}
\end{subfigure}
\caption{({\it The ultraviolet regime for UVB}) (a) In this paper, we are mainly interested in the regime of large $\sigma$ where one can informally view the model as being in the ``ultraviolet'' regime since samples from $Y=X+N(0,\sigma^2 I_d)$ (represented in violet in the figure) have (very) high energy in relation to samples from $X$ (represented in black). In addition\textemdash very related, but more important conceptually\textemdash the concentration of $Y$ is smoother and is higher dimensional than that of $X$ which the schematic is meant to capture. In the figure, the arrows represent $-\nabla f_\vartheta(y)$ which in UVB is parametrized by a VAE with parameters $\vartheta=(\theta,\phi)$, where the energy $f_\vartheta(y)$ is taken to be the ELBO computed by the VAE, but crucially the parameters are learned using the denoising machinery of DEEN with the Bayes estimator $\widehat{x}_\vartheta(y)=y-\sigma^2 \nabla f_\vartheta(y)$. The learning algorithm is to minimize $\Vert x- \widehat{x}_\vartheta(y) \Vert^2$ on expectation. (b) The VAE is set up such that the prior $p(z)=\mathcal{N}(z|0,I_{d_z})$ is the standard isotropic Gaussian. Assuming $d_z \gg 1$, $Z$ is approximately distributed according to $\text{Unif}(\sqrt{d_z} S_{d_z-1})$ which the schematic is meant to capture. }  \label{fig:symbolic}
 \end{center}
 \end{figure}

In neural empirical Bayes\textemdash referred to by DEEN due to its origin in~\citep{saremi2018deep}\footnote{``Deep Energy Estimator Networks'' was based prominently on denoising score matching~\citep{vincent2011connection}, itself rooted in \emph{score matching}~\citep{hyvarinen2005estimation}, and with connections to denoising autoencoders~\citep{alain2014regularized}; but as expressed in~\citep{saremi2019neural} we maintain the view that for the problem of approximating $\nabla \log p(y)$, \emph{empirical Bayes} is the more fundamental formulation. Also, regarding combining score matching and latent variable models, we should point out~\citep{swersky2011autoencoders, vertes2016learning} with fundamentally different approaches than here as they are both based on Hyv\"{a}rinen's score matching: they are formulated for learning the energy function of $X$\textemdash not $Y$\textemdash and therefore algorithmically they require the computation of second-order derivatives of the energy function.}\textemdash the learning problem is set up by parametrizing the energy function of $Y$ with a neural network with parameters $\vartheta$, denoted by $f_\vartheta: \mathbb{R}^d \rightarrow \mathbb{R}.$ We remind the reader that the energy function of $Y$ is defined as $-\log p(y)$ modulo an additive constant. With the energy function parametrization, the Bayes estimator (Eq.~\ref{eq:miyasawa}) takes the following form:
\< \label{eq:xhat} \widehat{x}_\vartheta(y) = y - \sigma^2 \nabla f_\vartheta(y).  \>
DEEN's learning algorithm is based on the following objective, minimized with SGD:
\<
	\label{eq:deen} \mathcal{L}(\vartheta) = \mathbb{E}_{x,y} \Vert x-\widehat{x}_\vartheta(y) \Vert^2, \\	
\>
where  the expectation is over the empirical distribution on $(x_i,y_{ij})$ pairs:
\< \label{eq:yij} y_{ij} = x_i+\varepsilon_j,\text{ where }\varepsilon_j \sim N(0,\sigma^2 I_d).\>
In summary, there are two main ``relaxations'' in formulating  density estimation in DEEN: (i) we are modeling the \emph{smoothed} density\footnote{The learned energy function has an implicit dependence on $\sigma$, the hyperparameter in $Y=X+N(0,\sigma^2 I_d)$. } that is associated with $Y,$ not $X$ (see Figure~\ref{fig:symbolic}a) (ii) the statistical model is \emph{unnormalized}~\citep{hyvarinen2005estimation}.

\subsection{Contributions}
\subsubsection{Big Picture}
DEEN appears to be a powerful and efficient learning framework for what it was designed to solve: learning unnormalized smoothed densities with the methodology of empirical Bayes. The ``power'' comes from the neural network parametrization~\citep{bengio2009learning, lu2017expressive} and the ``efficiency'' comes from SGD and the fact that the energy is computed \emph{deterministically}, short-cutting the inference step.  But can we do better?

Conceptually, one insight in~\citep{saremi2019neural} was to make a case in favor of probabilistic modeling of $Y=X+N(0,\sigma^2 I_d)$ instead of $X$. But what is missing there is  a \emph{latent space} where one can reason about noisy data. A natural question arises: can we set up the learning problem in terms of  the denoising methodology of empirical Bayes, but also have a latent space? This quest was conceptual/aesthetic at first\textemdash i.e. to make the model \emph{``more Bayesian''}\textemdash but to our surprise it also came with a significant boost to DEEN itself. 
\subsubsection{Learning Algorithm}
The sketch of the learning algorithm is as follows:
\begin{itemize}
	\item[(0.1)] Following up on~\citep{saremi2020learning}, we define a latent variable model for the random variable $Y$, where the joint density  $ p_\theta(y,z)$ is set up by taking the prior to be the isotropic Gaussian $p(z)=\mathcal{N}(z|0,I_{d_z})$ and taking the  conditional density to be \< \label{eq:intro-noise-model} p_\theta(y|z) = \mathcal{N}(y|\mu_y(z,\theta),\sigma^2 I_d).\>
	The evidence lower bound (ELBO) for $\log p_\theta(y)$ follows immediately:
	\< \label{eq:elbo-logpy-intro} \log p_\theta(y) \geq - \frac{1}{2\sigma^2}\mathbb{E}_{q_\phi(z|y)} \Vert y - \mu_y(z,\theta)\Vert^2 - \mathcal{D}_{\rm KL}[q_\phi(z|y),p(z)]-\log(2 \pi \sigma^2)^{d/2}. \>
	\item[(0.2)] \cite{saremi2020learning} started with the setup above with the goal of formulating Bayesian inference for very noisy data and they introduced the notion of \emph{imaginary noise model} defined by $p_\theta(x|z) = \mathcal{N}(x|\widehat{x}_\theta(z),\sigma^2 I_d) $  and studied ELBO for $\log p_\theta(x)$ instead. In short, in that framework $Y$ is out of the picture, i.e. the noisy data is not seen in training the model. The noise model was called ``imaginary'' since the measurement kernel $p(y|x)$ (for real noise) dictated how the latent variable model for the clean data was set up. In short, the idea was to \emph{implicitly} model noise with Bayesian inference.
	\item[{(1.1)}] Here we take a very different route where we aim at \emph{explicitly} modeling the noisy data with the noise model given in Eq.~\ref{eq:intro-noise-model}. The main idea is to bring back the empirical Bayes, but now we parametrize the energy function with the negative of the ELBO (modulo an additive constant). We therefore arrive at the following form for the energy function:
	\<f_\vartheta(y) = \frac{1}{2\sigma^2}\mathbb{E}_{q_\phi(z|y)} \Vert y - \mu_y(z,\theta)\Vert^2 + \mathcal{D}_{\rm KL}[q_\phi(z|y),p(z)].\>
Here $ \vartheta = (\theta,\phi)$.  
	\item[{(1.2)}] The negative ELBO is known as \emph{variational free energy}. It is called the energy function here since, crucially, instead of minimizing it (maximizing the ELBO), we follow~\citep{saremi2019neural} and set up the empirical Bayes learning objective:
	$$ \mathcal{L}(\vartheta) = \mathbb{E}_{(x,y)} \Vert x - \widehat{x}_\vartheta(y) \Vert^2,\text{ where } ~\widehat{x}_\vartheta(y) = y - \sigma^2 \nabla f_\vartheta(y).$$
\end{itemize}
This framework for approximating $\nabla \log p(y)$ with the unified machinery of empirical Bayes and variational Bayes is named \emph{unnormalized variational Bayes}: ``unnormalized'', since we only approximate $\log p(y)$ modulo an additive constant; and ``variational Bayes'', since it is indeed the workhorse for learning even though the learning objective itself is \emph{not} based on maximum likelihood estimation but based on the empirical Bayes least-squares estimation.
\subsubsection{Empirical Results}
Empirically, the following are the two highlights in the paper:
\begin{itemize}
	\item[(i)] One may postulate that nothing should be gained here empirically: ``nothing'', in terms of obtaining a lower loss. The end result is approximating $\nabla \log p(y)$ and if a neural network can do just that, why should one go through this lengthy program? But as it turns out\textemdash given equal ``capacity'' as defined by the dimension of $\vartheta$\textemdash there is a \emph{significant} gap in the loss between parametrizing $f(y)$ with a MLP (as it was done in DEEN) versus the parametrization in the variational Bayes machinery of UVB.
	\item[(ii)] We also revisit ``walk-jump sampling'' proposed in~\citep{saremi2019neural}, which we explore here for $\sigma=1$ (the random variable $X$ takes values in $[0,1]^d$). \emph{Why such a large $\sigma$?} We would like to explore a new paradigm in generative modeling, a middle ground between obtaining high-quality samples and having (very) fast-mixing MCMC samplers. 
	The walk-jump sampling is designed to be such a ``middle ground''. On MNIST, for $\sigma=1$, we demonstrate that in a single run\textemdash without a restart\textemdash we can traverse all classes of digits in a variety of styles with a fast-mixing sampler, sampling $Y$, which are mapped to $X$  by the jumps via $\widehat{x}_\vartheta(y)$ after every (only) 10 MCMC steps. \end{itemize}
\subsection{Outline}	
The remainder of the paper is organized as follows. In Sec.~\ref{sec:vae} we introduce variational autoencoders. In Sec.~\ref{sec:sigma-vae} we review $\sigma$-VAE which this work builds on. In Sec.~\ref{sec:uvb} we present our main contribution in unifying empirical Bayes and variational Bayes which, algorithmically speaking, brings together DEEN and $\sigma$-VAE. We then present the empirical results. In Sec.~\ref{sec:exp-architectures} we present the network architecture used for both DEEN and UVB chosen s.t. the dimensions of $\vartheta$ match. In Sections~\ref{sec:exp-loss} and~\ref{sec:exp-uvb} we present one of the most surprising result in the paper: UVB's significant improvements over DEEN accentuated for larger $\sigma$. In Sec.~\ref{sec:exp-scenicwalk} we present the walk-jump sampling results on MNIST for $\sigma=1$ to illustrate fast-mixing yet good quality sample generation capability of UVB in the very noisy regime. In Sec.~\ref{sec:uvb-vae} we look more closely at the encoder/decoder underlying the UVB, emphasizing UVB$\neq$VAE. In Sec.~\ref{sec:open} we discuss open questions, and we finish with a summary.

\clearpage

\section{Unnormalized Variational Bayes (UVB)} \label{sec:uvb}
Our plan for this section is to first introduce variational Bayes with a perspective suited for its \emph{unification} with empirical Bayes where one is interested in modeling $Y,$ not $X$. This paper also builds on~\citep{saremi2020learning} in formulating a notion of \emph{smoothed} variational inference which we briefly review. This is then followed by presenting our main contribution. 
\subsection{Auto-Encoding Variational Bayes} \label{sec:vae}
Consider the random variable $X$ in $\mathbb{R}^d$ in the context of \emph{latent variable models}, where we introduce the (latent) random variable $Z$ in $\mathbb{R}^{d_z}$ with a parametrized joint density $p_\theta(x,z)$ and our goal is to learn $\theta$ such that $p_\theta(x) = \int p_\theta(x,z) dz$ is a good approximation to $p(x)$. This is an intractable problem in general (in the absence of \emph{a priori} knowledge on the distribution of $X$) due to the curse of dimensionality. Taking Kullback--Leibler divergence $\mathcal{D}_{\rm KL}[p(x), p_\theta(x)]$  as the metric of choice to measure the approximation, and given the i.i.d. sequence $x_1,\dots,x_n$, the problem of learning $\theta$ is then equivalent to maximizing the log-likelihood: $ \mathcal{L}(\theta) = \sum_i \log p_\theta(x_i)$. In directed graphical models, one considers factorizing the joint density
$ p_\theta(x,z) = p_\theta(x|z) p_\theta(z)$, but the brute force maximum likelihood estimation of the parameters $\theta$ is in general intractable due to the computation 
$ \nabla_\theta \log p_\theta(x) =  \nabla_\theta \log \int p_\theta(x|z) p_\theta(z) dz$. 

In variational inference, one approaches this problem by studying yet \emph{another} intractable problem: approximating the posterior $ q_\phi(z|x) \approx p_\theta(z|x).$ Indeed, the posterior $p_\theta(z|x)$ itself is also intractable (in general) due to the Bayes rule $p_\theta(z|x)=p_\theta(x,z)/p_\theta(x)$, where the \emph{only} tractable density is the numerator $p_\theta(x,z)$ (assuming we have designed a tractable factorization). In other words, in probabilistic graphical models, the problem of modeling $X$ and the problem of the posterior inference over $Z$ are \emph{duals} and the complexity of the two problems ``match'' in some loose sense (see Remark~\ref{remark:duality} below). 

Taking a flexible yet tractable $q_\phi(z|x)$ as the candidate to approximate the posterior, it is straightforward to derive a lower bound for $\log p_\theta(x)$ using Jensen's inequality:
$$ \label{eq:elbo0} \log p_\theta(x) \geq \mathbb{E}_{q_\phi(z|x)} \log p_\theta(x|z) - \mathcal{D}_{\rm KL}[q_\phi(z|x), p_\theta(z)], $$
where $\mathcal{D}_{\rm KL}[q_\phi(z|x), p_\theta(z)]$ is the relative entropy between the approximate posterior and the prior, and the first term measures the reconstruction performance of the autoencoder. The right hand side in the inequality is referred to by evidence lower bound (ELBO) and the framework of choice for learning and inference is the variational autoencoder (VAE)~\citep{kingma2013auto,rezende2014stochastic} with the important invention of the \emph{reparameterization trick} that was developed to pass gradients through $z \sim q_\phi(z|x)$ crucial for  having low variance estimates of the gradients of the ELBO with respect to $\vartheta=(\theta,\phi)$.

\begin{remark} \label{remark:duality}The ``duality'' presented above is in fact another motivation for setting up a latent variable model for $Y$: since $Y$ is more tractable than $X$, therefore $p_\theta(z|y)$ must be more tractable than $p_\theta(z|x)$. More specifically, the standard assumptions made in the literature on (i) the Gaussian prior and (ii) the factorized Gaussian posterior, both appears to be very natural in studying $Y,$ and more so in higher dimensions and/or large $\sigma$ regimes. However, there are subtleties as we discuss next.  \end{remark}

\subsection{Imaginary Noise Models and $\sigma$-VAE} \label{sec:sigma-vae}
Now consider learning and inference in VAEs, where one is particularly interested in a formulation suited for $Y=X+N(0,\sigma^2 I_d)$ instead of $X$ (see Remark~\ref{remark:duality}). At a high level, this problem is also motivated by formalizing a notion of \emph{smoothed} variational inference as outlined in~\citep{saremi2020learning}, where the goal is to have a machinery to ``reason'' about noisy data and thus be \emph{robust to noise}, where smoothness and robustness are taken as duals, one implying the other.

On the surface, it seems straightforward to set up a latent variable model for the random variable $Y$, where $x$ in the presentation in the previous section needs to be changed to $y$. First, one needs to set up the joint density $p_\theta(y,z)=p_\theta(y|z)p_\theta(z)$. Due to the definition $Y=X+N(0,\sigma^2 I_d)$, a natural choice is given by $$p_\theta(y|z) = \mathcal{N}(y|\mu_y(z,\theta),\sigma^2 I_d),$$ together with the standard isotropic Gaussian prior (see Remark~\ref{remark:duality}). It is straightforward to derive the following lower bound for $\log p_\theta(y)$:
	$$ \log p_\theta(y) \geq - \frac{1}{2\sigma^2}\mathbb{E}_{q_\phi(z|y)} \Vert y - \mu_y(z,\theta)\Vert^2 - \mathcal{D}_{\rm KL}[q_\phi(z|y), p_\theta(z)], $$
where $-\log(2 \pi \sigma^2)^{d/2}$ is dropped from the r.h.s. as it does not affect the optimization of the ELBO ($\sigma$ is fixed).	 As stated in the introduction, we are especially interested in this problem in the regime of large noise (large $\sigma$). Just from the start, the variational inference formulated as such is ``doomed for failure'' as the stochastic estimators of the gradients of the ELBO will now have a high variance amplified by the Gaussian noise.  In contrast to the denoising machinery in neural empirical Bayes, here the clean data is completely left out of the picture. 

In~\citep{saremi2020learning}, they took a different route where they considered defining a notion of \emph{imaginary noise model}. In its summary, one starts with the real noise model as defined by $p(y|x)=\mathcal{N}(y|x,\sigma^2 I_d)$ and then use it as a \emph{template} for defining the noise model for the clean data:
\< \label{eq:imaginary-noise-model} p_\theta(x|z) = \mathcal{N}(x|\widehat{x}_\theta(z),\sigma^2 I_d),\>
 where the rationale for the notation $\widehat{x}_\theta(z)$ is that it is indeed the Bayes estimator of $X$ given $Z=z$: $ \widehat{x}(z,\theta)=\mathbb{E}[X|z]$. The model was named $\sigma$-VAE, summarized by:
\< \label{eq:sigma-vae-elbo} \log p_\theta(x) \geq  - \frac{1}{2\sigma^2}\mathbb{E}_{q_\phi(z|x)} \Vert x - \widehat{x}(z,\theta)\Vert^2 - \mathcal{D}_{\rm KL}[q_\phi(z|x),p_\theta(z)],\>
  After learning, \emph{if} the $\sigma$-VAE is shown noisy data, say sampled from $Y=X+N(0,\sigma^2 I_d)$\textemdash or any other noise processes\textemdash it first infers $z \sim q_\phi(z|y)$ which is then used to estimate $X$ by the  decoder $\widehat{x}_\theta(z)$. The $\sigma$-VAE was trained only seeing clean samples, but quite surprisingly, the imaginary noise model  was robust to large levels of (real) noise. Informally speaking, it appears that imaginary noise models are very expressive in dealing with noise,  but at the present time they cannot compete with DEEN with its denoising engine for learning. This motivated the model we present next that brings  DEEN and $\sigma$-VAE closer together.
  
\subsection{UVB = $\sigma$-VAE + DEEN} \label{sec:uvb}
In the space of models, neural empirical Bayes~\citep{saremi2019neural} (referred to by DEEN) and the $\sigma$-VAE~\citep{saremi2020learning} are very different as summarized below, where we highlight their key features both conceptually and from algorithmic standpoint:
\begin{itemize}
\item DEEN is based on the denoising machinery of empirical Bayes with the Bayes estimator of $X$ given by $\widehat{x}(y) = y+\sigma^2 \nabla \log p(y)$. The learning is formulated by parametrizing the negative $\log p(y)$ (modulo a constant) with a neural network $f_\vartheta: \mathbb{R}^d \rightarrow \mathbb{R}$ and the error signal $ \Vert x-y+\sigma^2 \nabla f_\vartheta(y) \Vert^2 $ is minimized on expectation using stochastic gradients. There are two types of noise present in the learning: \begin{itemize} \item the noise in the SGD, \item the noisy training data sampled from $p(y|x)$, \end{itemize} but \emph{there is no latent space and the model itself is fully deterministic}, e.g. given $y$, the model always returns the same result for $\widehat{x}_\vartheta(y)=y-\sigma^2 \nabla f_\vartheta(y)$. A key feature of DEEN inherited from empirical Bayes is that learning is not based on maximum likelihood estimation but based on \emph{least-squares estimation}.
\item $\sigma$-VAE is based on variational Bayes. There, the noisy data is \emph{absent} during learning and the model is purely based on maximizing the ELBO for $\log p_\theta(x)$ as stated in~(\ref{eq:sigma-vae-elbo}), but the model is structured such that the inference is smoothed and robust to noise, ``capable of denoising''. There are two types of noise present in learning: \begin{itemize}
\item the noise in the SGD,
\item the noise in inference $z \sim q_\phi(z|x).$
\end{itemize}
In contrast to DEEN, \emph{the latent space is at the heart of the learning framework}, e.g. given noisy data $y=x+\varepsilon$ at test time, the model infers $z \sim q_\phi(z|y)$ by the encoder and then returns $\widehat{x}_\theta(z)$ via the decoder. Finally, $\sigma$-VAE is based on \emph{maximum likelihood estimation} which is the starting point in the standard formulation of variational Bayes.
\end{itemize}

Here, our starting point is the same as~\citep{saremi2020learning} where we set up the latent variable model $p_\theta(y,z)=p_\theta(y|z)p_\theta(z)$ for $Y=X+N(0,\sigma^2 I_d)$ as defined by
\< \label{eq:noisemodel} p_\theta(y|z) = \mathcal{N}(y|\mu_y(z,\theta),\sigma^2 I_d),\>
where $\mu_y(z,\theta) \in \mathbb{R}^d$ is parameterized by a neural network and we take the prior to be \< \label{eq:prior} p(z)=\mathcal{N}(z|0,I_{d_z}).\> In this work, we take the approximate posterior to the factorized Gaussian (see Remark~\ref{remark:duality}):
\< \label{eq:approximate-posterior} q_\phi(z|y) = \prod_{k=1}^{d_z} \mathcal{N}(z_k|\mu_k(y,\phi),\sigma_k^2(y,\phi)),\>
where $k$ is the index for the dimensions in  the latent space $\mathcal{Z}=\mathbb{R}^{d_z}$. With this setup, one obtains the following lower bound for $\log p_\theta(y)$:
\< \label{eq:elbo-logpy} \log p_\theta(y) \geq - \frac{1}{2\sigma^2}\mathbb{E}_{q_\phi(z|y)} \Vert y - \mu_y(z,\theta)\Vert^2 - \mathcal{D}_{\rm KL}[q_\phi(z|y),p(z)].\>
With the choices (i) the Gaussian prior and (ii) the factorized Gaussian posterior, the KL divergence term is easily derived in closed form~\citep{kingma2013auto}:  
\< \label{eq:DKL} -2 \mathcal{D}_{\rm KL} = \sum_k \left( 1+\lambda_k(y,\phi) - \mu_k^2(y,\phi)-\exp(\lambda_k(y,\phi))\right),~ \text{where}~\lambda_k = 2 \log \sigma_k.\> 
The important invention in variational autoencoders was the reparameterization trick which was devised to pass gradients through the noise present in the inference network $z \sim q_\phi(z|y)$~\citep{kingma2013auto,rezende2014stochastic}. For the factorized Gaussian posterior~(\ref{eq:approximate-posterior}), the reparametrization takes the form:
\< \label{eq:reparameterization} z_k = \mu_k(y,\phi) + \sigma_k(y,\phi) \varepsilon_k,\text{ where } \varepsilon_k \sim N(0,1).\>

\begin{mainidea}
How can we bring empirical Bayes to the picture? The idea is somewhat straightforward conceptually. It is to use the VAE machinery to parametrize the energy function of $Y,$ identify it by the negative of the ELBO for $\log p_\theta(y)$. In the simple setup here, the energy function takes the form
\< \label{eq:fy} f_\vartheta(y) = \frac{1}{2\sigma^2}\mathbb{E}_{q_\phi(z|y)} \Vert y - \mu_y(z,\theta)\Vert^2 + \mathcal{D}_{\rm KL}[q_\phi(z|y),p(z)],\>
where $\vartheta=(\theta,\phi)$, and the relative entropy $\mathcal{D}_{\rm KL}$ is computed by Eq.~\ref{eq:DKL}. We then follow~\citep{saremi2019neural} is setting up the learning objective
\< \label{eq:learning-objective} \mathcal{L}(\vartheta) = \mathbb{E}_{(x,y)} \Vert x - y + \sigma^2 \nabla f_\vartheta(y) \Vert^2 \>
which is minimized with SGD. At optimality, $$ \nabla f_{\vartheta}(y) \approx -\nabla \log p(y).$$
\end{mainidea}
In plain words, we only use the VAE machinery to parameterize the energy function: the learning objective is not based on maximum likelihood estimation but based on the empirical Bayes least-squares estimation. This framework which is designed to approximate $\nabla \log p(y)$ with the unified machinery of empirical Bayes and variational Bayes is named unnormalized variational Bayes (UVB); ``unnormalized'', since we can only approximate $\log p(y)$ modulo an additive constant; ``variational Bayes'' since it is indeed the workhorse for learning the unnormalized density.
 
There are three types of noise present in the learning of the energy function:
\begin{itemize}
	\item[(i)] The noise in SGD,
	\item[(ii)] The (very) noisy training data $y$ sampled from $p(y|x) = \mathcal{N}(y|x,\sigma^2 I_d)$,
	\item[(iii)] The noise in the inference $z \sim q_\phi(z|y)$ sampled with the reparametrization trick~(\ref{eq:reparameterization}).
\end{itemize}
Informally speaking, empirical Bayes and variational Bayes play an ``equally'' important role here. From the perspective of learning dynamics, the presence of the three types of noise (described above) is to our knowledge novel, but also troubling at first: one may suspect that the model should not work at all! But the most surprising empirical result in the paper is that the new machinery significantly improves on~\citep{saremi2019neural} where they parametrized the energy function $f_\vartheta(y)$ with a very wide (highly overparametrized) ConvNet. As we demonstrate the significant quantitative improvements in the loss become qualitative, as visualized by the Bayes estimator $\widehat{x}_\vartheta(y)$ in the very noisy (ultraviolet) regime.

\section{Experiments} \label{sec:experiments}
\subsection{Network architecture} \label{sec:exp-architectures}
\subsubsection{UVB}  The encoder/decoder architecture here is the same one used in~\citep{saremi2020learning}. There are three neural networks in UVB: (i) $\lambda(\phi_1): \mathbb{R}^d \rightarrow \mathbb{R}^{d_z}$ (ii) $\mu(\phi_2): \mathbb{R}^d \rightarrow \mathbb{R}^{d_z}$, (iii) $\mu_y(\theta): \mathbb{R}^{d_z} \rightarrow \mathbb{R}^d$. The architectures for $\lambda$ and $\mu$ were the same, without any weight sharing. They were both ConvNets with  the expanding  channels=$(32,64,128)$, without pooling, and one fully connected layer with 200 neurons, and the linear readout with $d_z=100$. The decoder $\mu_y$ had one hidden layer with 2000 neurons and the logistic readout. Throughout, the activation function was $u \mapsto u/(1+\exp(-u))$, a smoothed-out ReLU that comes with at least two different names: ``SiLU''~\citep{elfwing2017sigmoid} and ``Swish''~\citep{ramachandran2017swish}. 

We chose the Adam optimizer~\citep{kingma2014adam} with the batch size of 128 and the constant learning rate of $10^{-4}$ trained for 400 epochs implemented in PyTorch~\citep{paszke2017automatic}. {\it The standard constant learning rate $\it 10^{-3}$ was not stable}, which we think it is due to the three types of noise present in the learning as discussed at the end of Section~\ref{sec:uvb}. We also ran some experiments with smaller batch size of 64 and 32 and the learning was stable arriving at similar losses after 400 epochs in the range of $\sigma$ reported here. 

In both UVB and DEEN (see below), the architectures for MNIST and CIFAR10 were the same. The \emph{architecture search} for this problem is important (especially in exploring large $d_z$ for complex distributions) but that is beyond the scope of this paper. One crucial choice both in UVB and DEEN is the activation function. The choice of a \emph{smooth} activation function is important here\footnote{With ReLU, the optimizer saturates at significantly higher loss (compared to SiLU/Swish).} since $\nabla_y f_\vartheta(y_{ij})$ must be computed first before computing the stochastic gradients $\nabla_\vartheta \widetilde{\mathcal{L}}(\vartheta)$ for updating parameters ($\widetilde{\mathcal{L}}$ is the loss evaluated on mini batches).

\subsubsection{DEEN}   
There is only one neural network in DEEN, $f(\vartheta):\mathbb{R}^d \rightarrow \mathbb{R}$, which parametrizes the energy function. We used a ConvNet with a similar architecture as in~\citep{saremi2019neural}, but smaller with the expanding channels=(128,256,512) and the fully connected layer of size 100. In Table~\ref{table:vartheta} we compare the dimensions of $\vartheta$ in DEEN vs. $\vartheta=(\theta,\phi_1,\phi_2)$ in UVB.

\begin{table}[h!]
\begin{center}
\begin{tabular}{c| c | c  }
\toprule
 dim($\vartheta$) & UVB & DEEN\\
\midrule 
MNIST & $2.7 \times 10^7$ & $2.6 \times 10^7$\\
CIFAR10 & $4.1 \times 10^7$ & $3.6 \times 10^7$\\
\bottomrule
\end{tabular}
\end{center}
\caption{ We report the dimensions of $\vartheta$ for UVB vs. DEEN in the experiments presented.   } 
\label{table:vartheta}
\end{table}


\subsection{UVB's capacity in approximating the score function $\nabla \log p(y)$ vs. DEEN's} \label{sec:exp-loss}
We start with UVB's improvements over DEEN where we report both the training and test losses at the end of training with constant learning rate as stated in previous section. The results are reported in Tables~\ref{table:mnist-loss} and~\ref{table:cifar-loss} which clearly demonstrates that UVB has a higher capacity in approximating $\nabla \log p(y)$ highlighted by the significant gaps in the losses between the two for larger noise levels. UVB's improvements over DEEN is an empirical observation and is not a fundamental result per se: DEEN is founded upon the \emph{universal approximation} capability of neural networks with its rich theory~\citep{cybenko1989approximation,hornik1989multilayer,lu2017expressive} and the end goal in both DEEN and UVB is approximating $\nabla \log p(y)$. Therefore \emph{if} the neural network $f_\vartheta: \mathbb{R}^d \rightarrow \mathbb{R}$ has large enough capacity one should be able to arrive at a solution such that $\nabla f_\vartheta(y)$ is arbitrarily close to $-\nabla \log p(y)$ pointwise. However, that is a big if! Ultimately, the neural network is finite with a finite capacity and the parametrization and the optimization both play important roles in learning. 

\begin{remark} \label{remark:gap} In Tables~\ref{table:mnist-loss} and~\ref{table:cifar-loss}, one also notices a clear qualitative difference between UVB and DEEN in terms of the generalization gap (in abuse of terminology, taken to be the difference between the test loss and the training loss). In DEEN, there is practically no difference between training and test losses while in UVB there is a gap which quite surprisingly persists to very high levels of noise. This gives yet another perspective on UVB's higher capacity in optimizing the empirical Bayes denoising loss.
\end{remark}

\begin{remark} For a sanity check, we also experimented with UVB with a Laplace decoder as outlined  in~\citep{saremi2020learning}, where we assumed (wrongly) that the noise model had been the Laplace distribution even though the training data was the Gaussian corrupted samples. Not surprisingly, the learning capacity of the model becomes very limited, where for $\sigma=1$ on MNIST the training loss saturated at $58.4$. This experiment validates our expectation that the design of the VAE itself plays an important role in learning the energy function $f_\vartheta(y)$.
\end{remark}
\vspace{0.1cm}

\begin{table}[h!]
\begin{center}
\begin{tabular}{c| c c c c c c c c }
\toprule
$\sigma$   & $\it 0.3$& $\it 0.4$& $\it 0.5$ & $\it 0.6$ & $\it 0.7$ & $\it 0.8$ & $\it 0.9$ & $\it 1.0$ \\
\midrule 
UVB (train)  & {\bf2.60} 	& {\bf3.84}  & {\bf5.09} & {\bf6.83} & {\bf8.72}  & {\bf10.6}  & {\bf12.6} & {\bf14.7} \\
DEEN (train)  & 2.77 	&  4.43  & 6.45& 8.87& 11.5 & 14.4  & 17.3 & 20.1 \\
\midrule 
UVB (test)  & 3.01 	& {\bf4.42}  & {\bf5.94} & {\bf7.55} & {\bf9.27}  & {\bf11.1} & {\bf13.0} &  {\bf15.0} \\
DEEN (test)  & {\bf2.89} 	& 4.48 & 6.47 & 8.83 & 11.5 & 14.3 & 17.1 & 19.9 \\
\bottomrule
\end{tabular}
\end{center}
\caption{ MNIST training/test losses: UVB vs. DEEN} 
\label{table:mnist-loss}
\end{table}

\begin{table}[h!]
\begin{center}
\begin{tabular}{c| c c c c c c c c }
\toprule
$\sigma$   & $\it 0.3$& $\it 0.4$& $\it 0.5$ & $\it 0.6$ & $\it 0.7$ & $\it 0.8$ & $\it 0.9$ & $\it 1.0$ \\
\midrule 
UVB (train)  & {\bf14.4} 	& {\bf20.1} & {\bf25.0} & {\bf29.7} & {\bf33.7}  & {\bf37.8} & {\bf42.1} & {\bf46.0}\\
DEEN (train) & 15.5 	& 21.5 & 27.1 & 32.6 & 37.9  & 43.0 & 48.1 & 52.6 \\
\midrule 
UVB (test)  &  {15.4} 	& {\bf20.7} & {\bf25.7} & {\bf30.5}  & {\bf34.7} & {\bf39.0} & {\bf42.9} & {\bf46.9} \\
DEEN (test)  &  15.5 	& 21.5  & 27.1 & 32.8 & 38.0 & 43.0 & 48.1 &  52.7 \\
\bottomrule
\end{tabular}
\end{center}
\caption{ CIFAR10 training/test losses: UVB vs. DEEN} 
\label{table:cifar-loss}
\end{table}

\subsection{The ultraviolet regime} \label{sec:exp-uvb}
We are especially interested in the regime of large $\sigma$. Here we present visualization experiments for $\sigma=1$ on MNIST~\citep{lecun1998gradient}. To have a sense of the scale $\sigma=1$, note that the largest $\ell_2$ distance in the hypercube $[0,1]^d$ is $\sqrt{d}$ the diagonal, i.e. $(0,0,\dots,0)$ to $(1,1,\dots,1)$, but the $\ell_2$ norm of the Gaussian noise $\varepsilon \sim N(0, I_d)$ itself concentrates at $\sqrt{d}$ for $d \gg 1$.\footnote{The geometric interpretation of noise level $\sigma$ with relation to pairwise distances in the dataset is discussed in~\citep{saremi2019neural} around the notion of ``$i$-sphere''. See Figure 2 in the reference.} The noise also happens to be quite high for \emph{our} visual system as illustrated in Figure~\ref{fig:mnistdenoise}.  Informally speaking, we call this the ultraviolet regime of noise where the noisy data become hardly recognizable.\footnote{Ultraviolet is defined at the onset of 750 THz in electromagnetic radiation, just above (in frequency) the violet range of the spectrum, which is invisible to humans although visible to many insects. Source: \url{https://en.wikipedia.org/wiki/ultraviolet}} The empirical Bayes estimator of $X$ given noisy samples $Y=x_i+\varepsilon,~\varepsilon\sim N(0,\sigma^2 I_d)$  for randomly selected samples from the MNIST test set is presented in Figure~\ref{fig:mnistdenoise} after learning the energy function $f_\vartheta$ for both UVB and DEEN. 

\subsubsection{Poor man's NEBULA}\label{sec:exp-nebula}\cite{saremi2019neural} defined a probabilistic notion of associated memory called NEBULA as attractors (strict local minima) of the energy function where the attractor dynamics is dictated by the gradient flow of the leaned energy $f_\vartheta$. We adopt the same definition here, where now the energy function is learned by UVB. Strictly speaking, to arrive at the attractors one should run the continuous gradient flow dynamics which in practice is simulated with gradient descent with small step sizes. In this work, we experimented with a ``poor man's'' version of NEBULA where we assume we only have the budget to take \emph{two} steps and  set the step size to be $\sigma^2$.  The results are reported in Figure~\ref{fig:mnistdenoise}d: if we were to take more steps, the denoised signals would have been cleaned up even more but that comes with unexpected consequences, changing the style of digits to more prototypical modes and the digit \emph{classes} themselves could also change.

\subsubsection{Failures of Vanilla (Gaussian) VAE in modeling $Y$} \label{sec:exp-failures}
Below we report the failures of the vanilla Gaussian VAE in modeling $Y=X+N(0,\sigma^2 I_d)$ for high levels of noise ($\sigma=1$). Here, the learning objective is to maximize the ELBO given in the r.h.s. of~(\ref{eq:elbo-logpy}). In the figure below, the top row are the clean samples $x_i$ from MNIST test set, the second row are noisy samples $y_i=x_i+\varepsilon$ where $ \varepsilon \sim N(0,I_d)$, and the third row are the corresponding $\mu_y(z,\theta)$, where $z \sim q_\phi(z|y_i)$. For completeness, in the last row we have also given the samples from the decoder $z \mapsto \mu_y(z,\theta)$, where $z \sim N(0,I_{d_z})$. 

  \begin{figure}[h!]
\begin{center}
  {\includegraphics[width=0.9\textwidth]{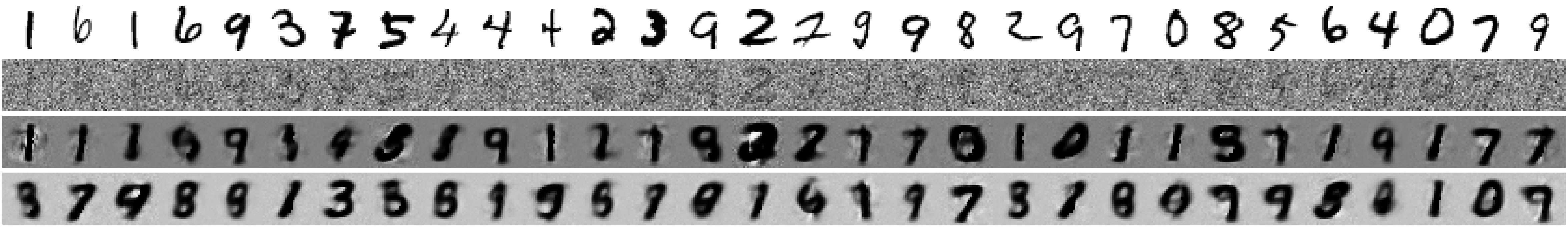}}
\end{center}
 \end{figure}

 \begin{figure}[b!]
 \begin{center} 
   \begin{subfigure}[$x_i$ from the MNIST test set]{\includegraphics[width=\textwidth]{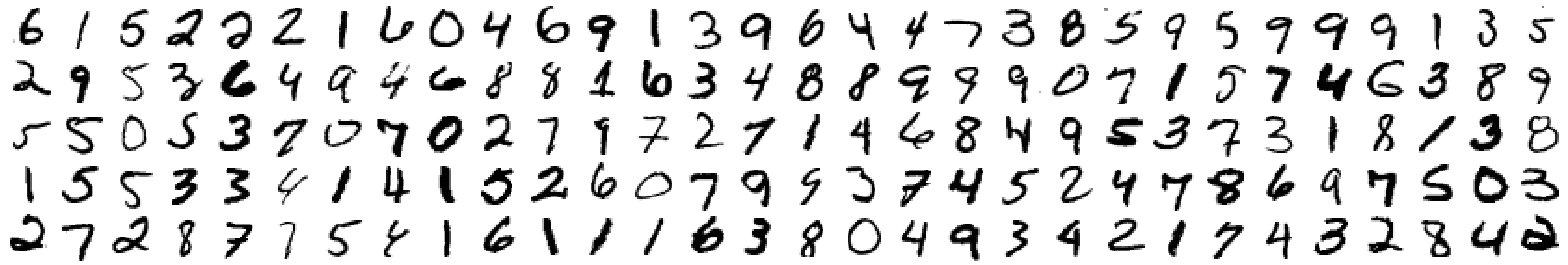}}
  \end{subfigure} 
     \begin{subfigure}[$y_{i} =x_i+\varepsilon$,~$\varepsilon \sim N(0,\sigma^2 I_d)$, where $\sigma=1$ ]{\includegraphics[width=\textwidth]{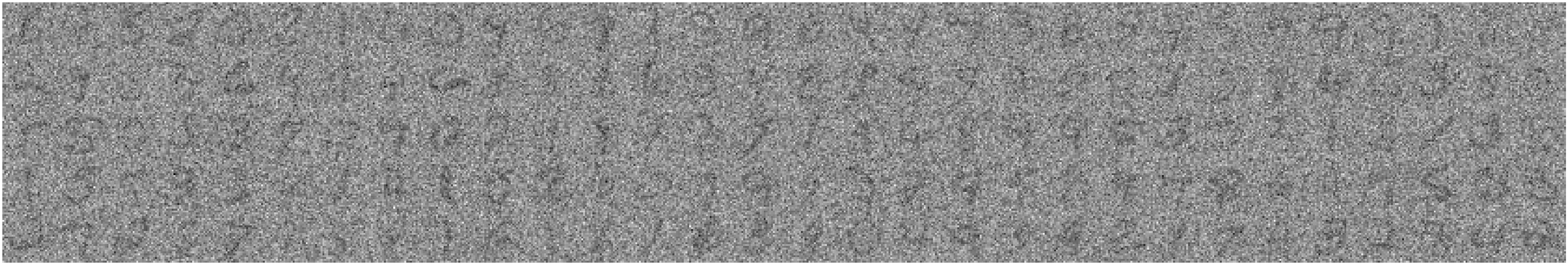}}
  \end{subfigure} 
     \begin{subfigure}[$\widehat{x}(y_{i})=y_{i}-\sigma^2 \nabla f_\vartheta(y_{i})$, where $f_\vartheta$ is learned with UVB for $\sigma=1$, test loss = 15.0]{\includegraphics[width=\textwidth]{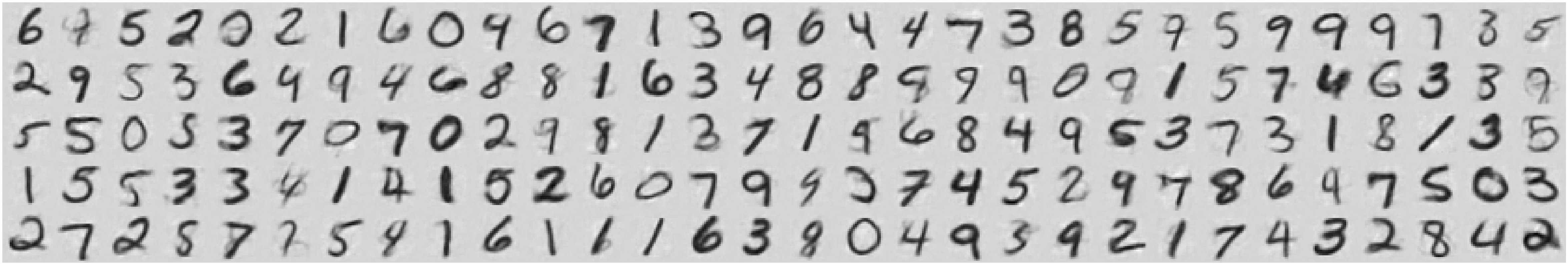}}
  \end{subfigure} 
     \begin{subfigure}[$\widehat{x}(\widehat{x}(y_{i})) = \widehat{x}(y_{i})-\sigma^2 \nabla f_\vartheta(\widehat{x}(y_{i}))$, where $f_\vartheta$ is the same as above ]{\includegraphics[width=\textwidth]{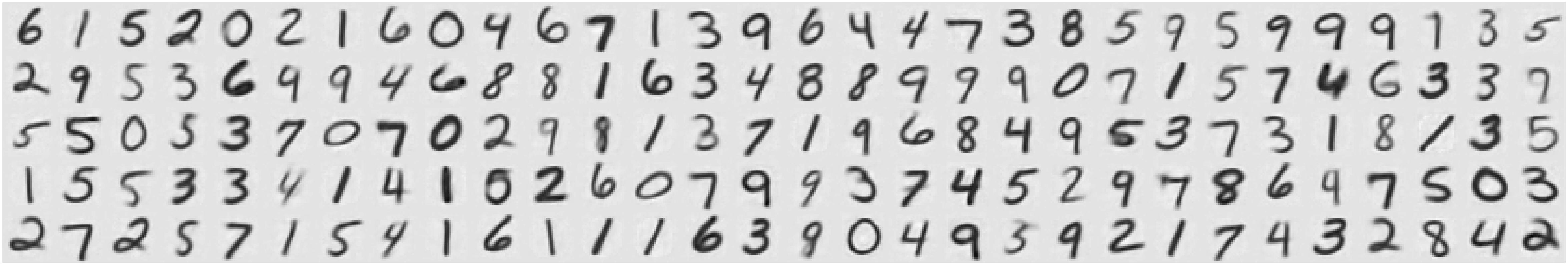}}
  \end{subfigure}       
   \begin{subfigure}[$\widehat{x}(y_{i})=y_{i}-\sigma^2 \nabla f_\vartheta(y_{i})$, where $f_\vartheta$ is learned with DEEN for $\sigma=1$, test loss = 19.9]{\includegraphics[width=\textwidth]{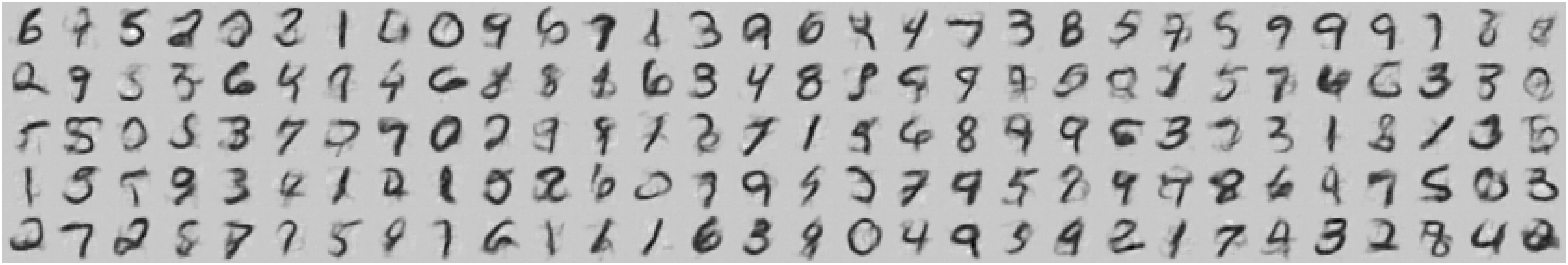}}
  \end{subfigure} 
 \caption{ {\it (denoising in the ultraviolet regime: UVB vs. DEEN)} (a) Clean samples. (b) Gaussian noise $\varepsilon$ is added to $x_i$ and it is held fixed. (c) The Bayes estimator is computed after learning the energy function $f_\vartheta(y)$ with UVB. Here, we only sampled $z$ once to compute the energy function. In principle, $\nabla f_\vartheta(y)$ must be computed by taking the expectation over $q_\phi(z|y)$. We did those experiments as well where we took $1000$ samples. However, the differences are marginal, bounded in [-0.0068,0.0068] for the set presented here.  (d) This experiment is discussed in length in Sec.~\ref{sec:exp-nebula}. (e) The Bayes estimator of $X$ computed by DEEN.  }
 \label{fig:mnistdenoise}
 \end{center}
 \end{figure}

  \begin{figure}[h!]
  {\includegraphics[width=\textwidth]{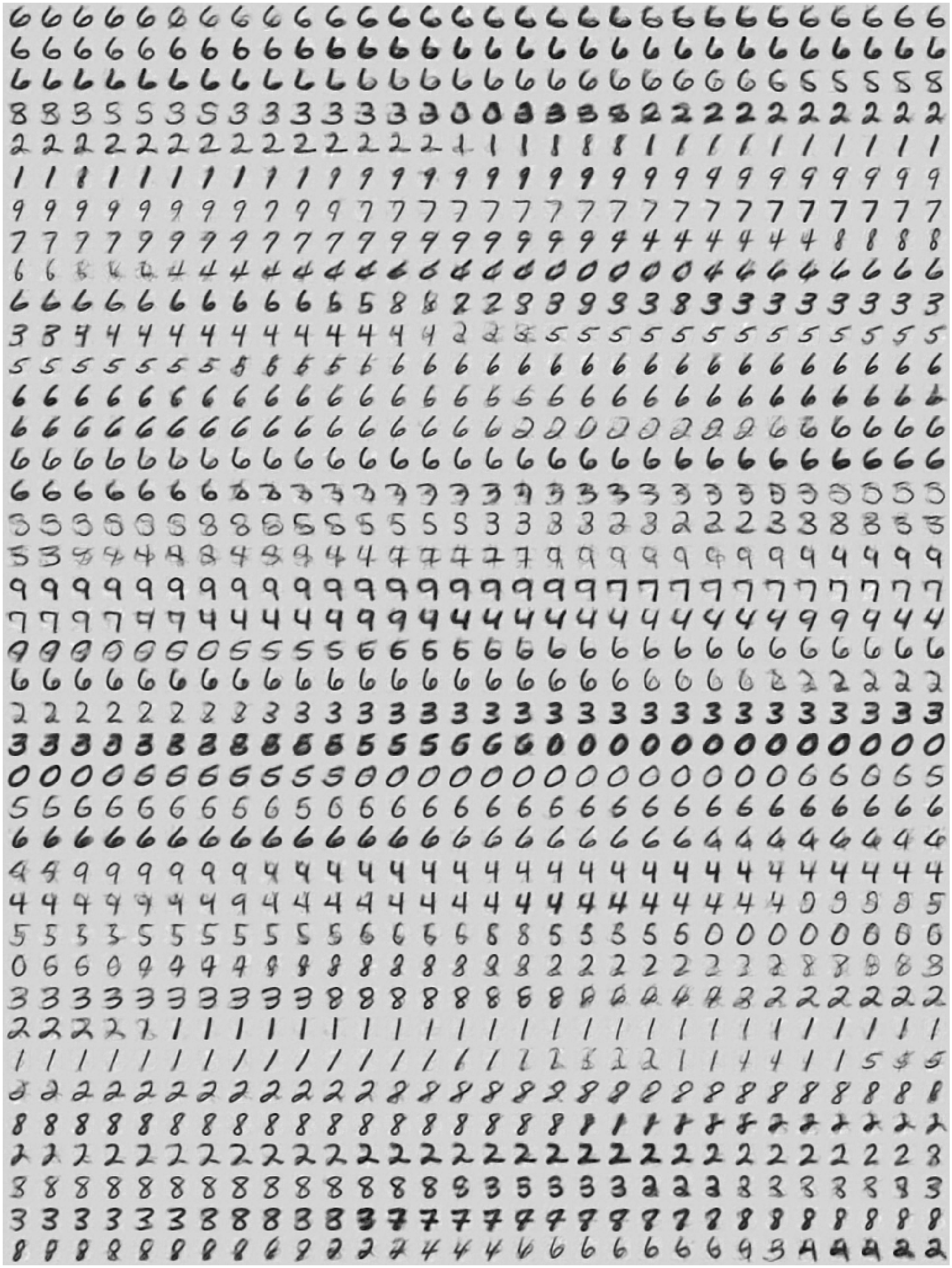}}
\caption{({\it a short scenic walk}) Walk-jump sampling  results where each sample is generated after just \emph{ten} Langevin MCMC steps and one empirical Bayes jump for $\sigma=1$. }
\label{fig:scenic-walk}
 \end{figure}

 \clearpage
 
\subsection{Fast-mixing walk-jump sampling in the ultraviolet regime} \label{sec:exp-scenicwalk}
What we have established so far empirically is that the \emph{unnormalized variational Bayes} (UVB) appears to have a much higher capacity (as measured by the loss obtained) in learning unnormalized densities than \emph{neural empirical Bayes} (DEEN). Due to the significant improvements of UVB over DEEN\textemdash especially, in the high noise regime\textemdash we revisited the walk-jump sampling introduced in~\citep{saremi2019neural} for $\sigma=1$. Walk-jump sampling finds a ``happy medium'' between the desire to generate sharp (high-quality) samples and the quest for MCMC to mix faster.  The algorithm is quite simple:
\begin{itemize}
	\item[(i)] Sample $Y$ after learning the energy function $f_\vartheta(y)$. Here as in~\citep{saremi2019neural} we consider (overdamped) Langevin MCMC:
	\< \label{eq:langevin} y_{t+1} = y_t - \delta^2 \nabla f_\vartheta(y_t) + \sqrt{2} \delta \varepsilon,\>
	where $\varepsilon \sim N(0,I_d)$, $\delta$ is the step size, and $t$ is the discrete time. One typically starts the MCMC at $y_0 \sim \text{Unif}([0,1]^d)$. In very noisy regime (and/or in very high dimensions) we can visualize~(\ref{eq:langevin}) as walking on/in the high-dimensional ``ultraviolet manifold'' where $Y=X+N(0,\sigma^2 I_d)$ is concentrated (see Figure~\ref{fig:symbolic}a).
	\item[(ii)]	 At \emph{arbitrary} times $\tau,$ use the Bayes estimator of $X$ to jump from $y_\tau$ to $\widehat{x}_\vartheta(y_\tau).$ In comparison with the Langevin MCMC, these samples are not statistical samples (taken from $X$) but going back to the definition of the Bayes estimator, they are the mean over the posterior:
	\< \label{eq:jump} \widehat{x}_\vartheta(y_\tau) = \int x p_\vartheta(x|y_\tau) dx .\>
	Essentially, with the jumps we get much sharper ``denoised'' samples than the noisy samples generated by the Langevin MCMC (if our goal is to see sharper samples). 
\end{itemize}

As it is clear above, walk-jump sampling finds a compromise between having fast-mixing MCMC samplers  (more on that in Remark~\ref{remark:time-complexity} below) and obtaining sharp samples ``close to'' data manifold where the random variable $X$ is concentrated. {\it From our perspective, the first quest is the more important one, i.e. in this paradigm, one is easily willing to sacrifice sample quality for faster mixing (see Remark~\ref{remark:hd}).} On MNIST, we present results for $\sigma=1$. On could do experiments with smaller $\sigma$ but $\sigma=1$ was chosen due to the (visually) good denoising performance of UVB presented in Figure~\ref{fig:mnistdenoise}c and to push the model to its limits, informally speaking bringing the model close to the schematic in Figure~\ref{fig:symbolic}a. 

The results are presented in Fig.~\ref{fig:scenic-walk}, where we set the step size $\delta=0.2$ (see Eq.~\ref{eq:langevin}) and demonstrate jump samples (Eq.~\ref{eq:jump}) after every 10 Langevin MCMC steps. There was no warmup phase in the experiment presented, i.e. the first sample shown is obtained after only 10 MCMC steps (the samples are viewed top-left to bottom-right). The experiment demonstrates the fast-mixing property of the Langevin MCMC in the ultraviolet regime while retaining some sharpness in the jump steps. In addition, samples shown in Fig.~\ref{fig:scenic-walk} can become sharper (cleaned up more) if they are followed up by another jump $\widehat{x}_\vartheta(\widehat{x}_\vartheta(y))$ as described in Sec.~\ref{sec:exp-nebula} and visualized in Fig.~\ref{fig:mnistdenoise}d but here we opted for the less sharp version here to highlight the transitions between the digits in the Langevin MCMC more clearly.

\begin{remark} \label{remark:time-complexity} There have been recent developments in analyzing Langevin MCMC for smooth and strongly convex energy functions~\citep{cheng2018underdamped}, and related results on statistical models where the energy function is not convex globally but $m$-strongly convex outside a ball of radius $R$~\citep{cheng2018sharp}. In the latter non-convex case (relevant here), the assumption is that there exists $m, R>0$ such that for all $y, y' \in \mathbb{R}^d$:
\< \langle \nabla f_\vartheta(y) - \nabla f_\vartheta(y'),y-y'\rangle \geq m \Vert y-y'\Vert^2\ ~ {\rm if}\ ~ \Vert y - y' \Vert_2>R.\>
As expected, one gains in time complexity in the convergence of Langevin MCMC to the true distribution for models that satisfy the condition above for smaller $R$.  On the other hand, it is clear that larger $\sigma$ lead to lower time complexity. Unfortunately, it is not tractable to analytically bound $m,R$ for $f_\vartheta$ parameterized by UVB, but assuming $f_\vartheta(y) \approx f(y)$, it is an open question whether one can analyze this problem for $f(y)$ and study  the role of $\sigma$ in the time complexity analysis.
\end{remark}

\begin{remark}\label{remark:hd}
	In this presentation we emphasized the regime of interest on having both fast-mixing samplers and good quality samples\textemdash ``good enough,'' that is! But finding that middle ground is clearly dataset and application dependent. On that note, generative models are sometimes motivated by the ability of humans in dreaming or in planning for actions, but what is usually left out is that we (humans) typically do not plan/dream in high resolution. On the other hand, stretching the (informal) analogy between planning/dreaming and sampling probabilistic models further, we appear to have very fast-mixing samplers built in our brain.
	\end{remark}

\subsection{UVB $\neq$ VAE} \label{sec:uvb-vae}
In this section, we look more closely at UVB ``under the hood''. In particular, we show empirical results on the fact that the encoder/decoder underlying the UVB learned by least-squares estimation is very different than the one learned by maximum likelihood estimation: symbolically, UVB $\neq$ VAE.  This must be clear since the learning paradigms are very different but it is worth to be emphasized through experiments by looking at both the encoder and the decoder of the VAE learned by UVB's least-squares objective and comparing it with the one learned by maximizing the ELBO as discussed in Sec.~\ref{sec:exp-failures}.

\subsubsection{UVB's encoder}
We first report $\mathcal{D}_{\rm KL}[q_\phi(z|y),p(z)]$ of the encoder of the UVB for $\sigma=1$ and compare it with $\mathcal{D}_{\rm KL}$ learned by maximizing the ELBO discussed in Sec.~\ref{sec:exp-failures}. The encoder-decoder architectures are the same; the learning objectives, vastly different! The results are given in Table~\ref{table:DKL} showing the astronomical 688 nats differences between the two.

\begin{table}[h!]
\begin{center}
\begin{tabular}{c| c | c  }
\toprule
$\sigma=1$ & UVB & VAE\\
\midrule 
$\langle \mathcal{D}_{\rm KL}[q_\phi(z|y),p(z)] \rangle$ & $692$  & $3.70$\\
\bottomrule
\end{tabular}
\end{center}
\caption{ The encoder of the trained models (UVB vs. VAE) on MNIST. } 
\label{table:DKL}
\end{table}

\subsubsection{UVB's decoder}
This section is the repeat of the experiments of the previous section but here we compare the reconstruction term $\langle \mathbb{E}_{q_{\phi}(z|x)} \Vert y-\mu_y(z,\theta)  \Vert^2 \rangle/(2 \sigma^2)$ between the two models, where $\langle \cdot \rangle$ is evaluated on the test set. The results are reported in Table~\ref{table:reconstruction}.

\begin{table}[h!]
\begin{center}
\begin{tabular}{c| c | c  }
\toprule
$\sigma=1$ & UVB & VAE\\
\midrule 
$\langle \mathbb{E}_{q_{\phi}(z|x)} \Vert y-\mu_y(z,\theta)  \Vert^2 \rangle /(2 \sigma^2) $ & $103$  & $51.5$\\
\bottomrule
\end{tabular}
\end{center}
\caption{ The reconstruction term for the two trained models on MNIST on the test set. } 
\label{table:reconstruction}
\end{table}

\begin{figure}[h!] 
\begin{center}
\begin{subfigure}[Samples $\mu_y(z,\theta),~z\sim N(0,I_{d_z})$ from the decoder of UVB for a trained model for MNIST for $\sigma=1$]
 {\includegraphics[width=\textwidth]{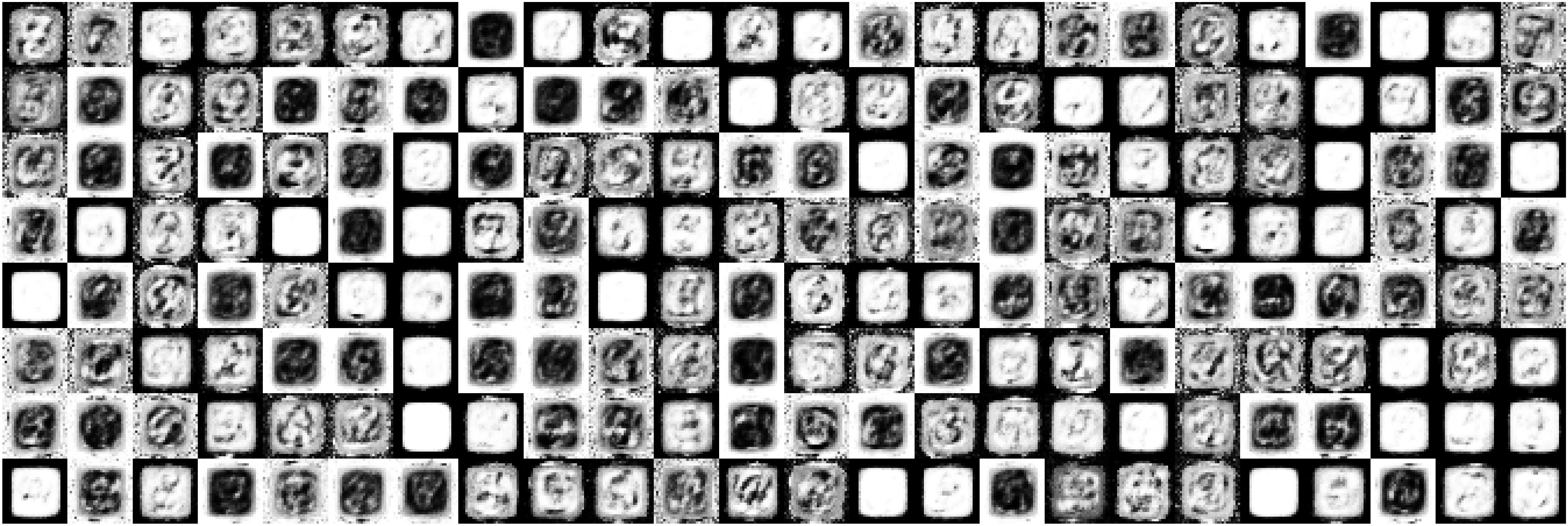}}
\end{subfigure}
\begin{subfigure}[Samples $\mu_y(z,\theta),~z\sim N(0,I_{d_z})$ from the decoder of UVV for a trained model for CIFAR10 for $\sigma=1$]
 {\includegraphics[width=\textwidth]{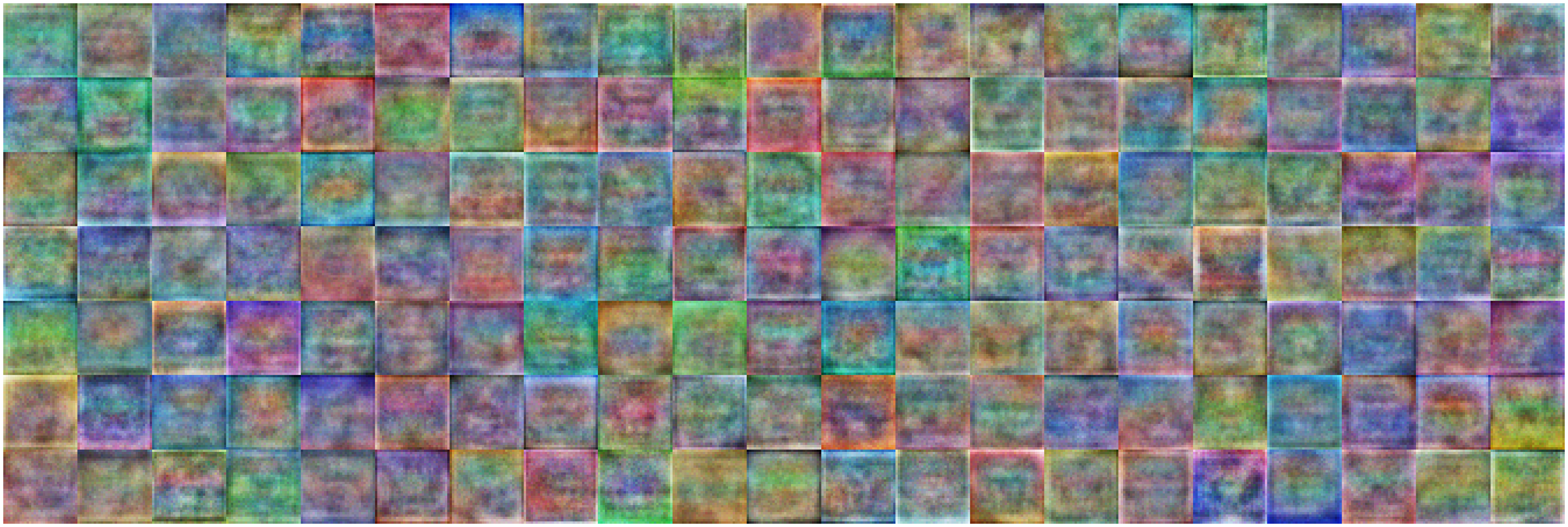}}
\end{subfigure}
\caption{({\it UVB$\neq$VAE}) In this figure, we visualize the samples generated by the decoder underlying the UVB for the trained models on MNIST and CIFAR10 for $\sigma=1$.  }  \label{fig:decoder}
 \end{center}
 \end{figure}

Note that neither the reconstruction term nor the KL divergence term themselves enter the learning objective in UVB: \emph{only their gradients}. To shed more light on this, in Figure~\ref{fig:decoder} we have also included visualizations of the decoder given by
$ z\mapsto \mu_y(z,\theta),~\text{where}~ z \sim N(0,I_{d_z})$
 for both MNIST and CIFAR10, giving yet another perspective on the fact that UVB$\neq$VAE.

\section{Open Questions} \label{sec:open}
We highlight two open questions:
\begin{quote}
	{\it 1. Why does UVB have a higher capacity than DEEN for approximating $\nabla \log p(y)$?}
\end{quote}
 This issue can be looked at from the perspective of implicit parametrization. Note that in both DEEN and UVB the learning objective is set up to approximate the score function $\nabla \log p(y) \in \mathbb{R}^d $, but in both cases, we parametrized the energy function $f_\vartheta(y) \in \mathbb{R}$ instead. One could view energy function as \emph{implicitly} parametrizing the score function; ``implicit'', since $\nabla f_\vartheta(y)$ is computed with automatic differentiation and (in practice) its symbolic form is not known~\citep{griewank2003mathematical, baydin2017automatic}. This implicit parametrization view of DEEN for approximating the score function was discussed in~\citep{saremi2019approximating}. In UVB this implicit parameterization is taken to a ``higher level'' where the energy function \emph{itself} is not parametrized by a MLP but by the ELBO computed by a VAE. 
 
 \begin{quote}
	{\it 2. Why does the generalization gap in UVB persist to very high levels of noise?}
\end{quote}
In DEEN, there is no distinction between the test set and the training set for higher levels of noise: on CIFAR10 starting at $\sigma=0.3$ and on MNIST starting at $\sigma=0.4$. However, for UVB a gap persists all the way to $\sigma=1$\textemdash we should point out that the test loss in our experiments did not go up during training for the range of $\sigma$ reported here, and the generalization gap is not a sign of overfitting. (In general, in this denoising framework, it must be very difficult to overfit for large noise levels.)  For CIFAR10, it is a next step to repeat the experiments here on the 80 million tiny image database~\citep{torralba200880}\footnote{It appears that the database have been taken down at the present time.} which CIFAR10~\citep{krizhevsky2009learning} is a small subset of and check whether this gap closes.

\section{Summary}
We introduced unnormalized variational Bayes (UVB) as a unification of empirical Bayes and variational Bayes for approximating unnormalized densities. Algorithmically, the energy function was parameterized and computed by a variational autoencoder but the learning itself was formulated via empirical Bayes. What did we gain in this ``unification''? In variational Bayes, one should be mindful of how tight the evidence lower bound (ELBO) is; there are no such worries here since the learned density is \emph{unnormalized}. In addition, as we demonstrated empirically, (to our surprise) the learning itself is boosted significantly by the variational autoencoder (VAE) parametrization. At the present time, it is difficult to tease apart whether this ``boost'' is mostly due to the parametrization or the fact the optimization is more regularized due to the inner workings of \emph{variational Bayes}. In this paper, the VAE was set up by an isotropic Gaussian prior, a factorized Gaussian posterior (for the encoder), and a Gaussian conditional density (for the decoder), but this is just one instantiation of UVB and more expressive priors/encoders/decoders can be used where the only overhead to have in mind is the computation of the \emph{gradient} of the ELBO with respect to noisy inputs.

\section*{Acknowledgement} 
I am grateful to Aapo Hyv\"{a}rinen and Francis Bach for their comments on the manuscript and to my colleagues Christian Osendorfer and Rupesh Srivastava for discussions. 

\bibliography{uvb.bib}

\begin{thebibliography}{30}
\providecommand{\natexlab}[1]{#1}
\providecommand{\url}[1]{\texttt{#1}}
\expandafter\ifx\csname urlstyle\endcsname\relax
  \providecommand{\doi}[1]{doi: #1}\else
  \providecommand{\doi}{doi: \begingroup \urlstyle{rm}\Url}\fi

\bibitem[Alain and Bengio(2014)]{alain2014regularized}
Guillaume Alain and Yoshua Bengio.
\newblock What regularized auto-encoders learn from the data-generating
  distribution.
\newblock \emph{Journal of Machine Learning Research}, 15\penalty0
  (1):\penalty0 3563--3593, 2014.

\bibitem[Baydin et~al.(2017)Baydin, Pearlmutter, Radul, and
  Siskind]{baydin2017automatic}
At{\i}l{\i}m~G{\"u}nes Baydin, Barak~A Pearlmutter, Alexey~Andreyevich Radul,
  and Jeffrey~Mark Siskind.
\newblock Automatic differentiation in machine learning: a survey.
\newblock \emph{The Journal of Machine Learning Research}, 18\penalty0
  (1):\penalty0 5595--5637, 2017.

\bibitem[Bengio(2009)]{bengio2009learning}
Yoshua Bengio.
\newblock Learning deep architectures for {AI}.
\newblock \emph{Foundations and Trends in Machine Learning}, 2\penalty0
  (1):\penalty0 1--127, 2009.

\bibitem[Cheng et~al.(2018{\natexlab{a}})Cheng, Chatterji, Abbasi-Yadkori,
  Bartlett, and Jordan]{cheng2018sharp}
Xiang Cheng, Niladri~S Chatterji, Yasin Abbasi-Yadkori, Peter~L Bartlett, and
  Michael~I Jordan.
\newblock Sharp convergence rates for {L}angevin dynamics in the nonconvex
  setting.
\newblock \emph{arXiv preprint arXiv:1805.01648}, 2018{\natexlab{a}}.

\bibitem[Cheng et~al.(2018{\natexlab{b}})Cheng, Chatterji, Bartlett, and
  Jordan]{cheng2018underdamped}
Xiang Cheng, Niladri~S Chatterji, Peter~L Bartlett, and Michael~I Jordan.
\newblock Underdamped {L}angevin {MCMC}: A non-asymptotic analysis.
\newblock In \emph{Conference on Learning Theory}, pages 300--323,
  2018{\natexlab{b}}.

\bibitem[Cybenko(1989)]{cybenko1989approximation}
George Cybenko.
\newblock Approximation by superpositions of a sigmoidal function.
\newblock \emph{Mathematics of control, signals and systems}, 2\penalty0
  (4):\penalty0 303--314, 1989.

\bibitem[Elfwing et~al.(2017)Elfwing, Uchibe, and Doya]{elfwing2017sigmoid}
Stefan Elfwing, Eiji Uchibe, and Kenji Doya.
\newblock Sigmoid-weighted linear units for neural network function
  approximation in reinforcement learning.
\newblock \emph{arXiv preprint arXiv:1702.03118}, 2017.

\bibitem[Griewank(2003)]{griewank2003mathematical}
Andreas Griewank.
\newblock A mathematical view of automatic differentiation.
\newblock \emph{Acta Numerica}, 12\penalty0 (1):\penalty0 321--398, 2003.

\bibitem[Hornik et~al.(1989)Hornik, Stinchcombe, and
  White]{hornik1989multilayer}
Kurt Hornik, Maxwell Stinchcombe, and Halbert White.
\newblock Multilayer feedforward networks are universal approximators.
\newblock \emph{Neural networks}, 2\penalty0 (5):\penalty0 359--366, 1989.

\bibitem[Hyv{\"a}rinen(2005)]{hyvarinen2005estimation}
Aapo Hyv{\"a}rinen.
\newblock Estimation of non-normalized statistical models by score matching.
\newblock \emph{Journal of Machine Learning Research}, 6\penalty0
  (Apr):\penalty0 695--709, 2005.

\bibitem[Jordan et~al.(1999)Jordan, Ghahramani, Jaakkola, and
  Saul]{jordan1999introduction}
Michael~I Jordan, Zoubin Ghahramani, Tommi~S Jaakkola, and Lawrence~K Saul.
\newblock An introduction to variational methods for graphical models.
\newblock \emph{Machine learning}, 37\penalty0 (2):\penalty0 183--233, 1999.

\bibitem[Kingma and Ba(2014)]{kingma2014adam}
Diederik~P Kingma and Jimmy Ba.
\newblock Adam: A method for stochastic optimization.
\newblock \emph{arXiv preprint arXiv:1412.6980}, 2014.

\bibitem[Kingma and Welling(2013)]{kingma2013auto}
Diederik~P Kingma and Max Welling.
\newblock Auto-encoding variational {B}ayes.
\newblock \emph{arXiv preprint arXiv:1312.6114}, 2013.

\bibitem[Krizhevsky(2009)]{krizhevsky2009learning}
Alex Krizhevsky.
\newblock Learning multiple layers of features from tiny images.
\newblock 2009.

\bibitem[LeCun et~al.(1998)LeCun, Bottou, Bengio, and
  Haffner]{lecun1998gradient}
Yann LeCun, L{\'e}on Bottou, Yoshua Bengio, and Patrick Haffner.
\newblock Gradient-based learning applied to document recognition.
\newblock \emph{Proceedings of the IEEE}, 86\penalty0 (11):\penalty0
  2278--2324, 1998.

\bibitem[Lu et~al.(2017)Lu, Pu, Wang, Hu, and Wang]{lu2017expressive}
Zhou Lu, Hongming Pu, Feicheng Wang, Zhiqiang Hu, and Liwei Wang.
\newblock The expressive power of neural networks: A view from the width.
\newblock In \emph{Advances in Neural Information Processing Systems}, pages
  6231--6239, 2017.

\bibitem[Miyasawa(1961)]{miyasawa1961empirical}
Koichi Miyasawa.
\newblock An empirical {B}ayes estimator of the mean of a normal population.
\newblock \emph{Bulletin of the International Statistical Institute},
  38\penalty0 (4):\penalty0 181--188, 1961.

\bibitem[Paszke et~al.(2017)Paszke, Gross, Chintala, Chanan, Yang, DeVito, Lin,
  Desmaison, Antiga, and Lerer]{paszke2017automatic}
Adam Paszke, Sam Gross, Soumith Chintala, Gregory Chanan, Edward Yang, Zachary
  DeVito, Zeming Lin, Alban Desmaison, Luca Antiga, and Adam Lerer.
\newblock Automatic differentiation in {PyTorch}.
\newblock 2017.

\bibitem[Ramachandran et~al.(2017)Ramachandran, Zoph, and
  Le]{ramachandran2017swish}
Prajit Ramachandran, Barret Zoph, and Quoc~V Le.
\newblock Swish: a self-gated activation function.
\newblock \emph{arXiv preprint arXiv:1710.05941}, 7, 2017.

\bibitem[Raphan and Simoncelli(2011)]{raphan2011least}
Martin Raphan and Eero~P Simoncelli.
\newblock Least squares estimation without priors or supervision.
\newblock \emph{Neural computation}, 23\penalty0 (2):\penalty0 374--420, 2011.

\bibitem[Rezende et~al.(2014)Rezende, Mohamed, and
  Wierstra]{rezende2014stochastic}
Danilo~Jimenez Rezende, Shakir Mohamed, and Daan Wierstra.
\newblock Stochastic backpropagation and approximate inference in deep
  generative models.
\newblock \emph{arXiv preprint arXiv:1401.4082}, 2014.

\bibitem[Robbins(1956)]{robbins1956empirical}
Herbert Robbins.
\newblock An empirical {B}ayes approach to statistics.
\newblock In \emph{Proc. Third Berkeley Symp.}, volume~1, pages 157--163, 1956.

\bibitem[Saremi(2019)]{saremi2019approximating}
Saeed Saremi.
\newblock On approximating $\nabla f $ with neural networks.
\newblock \emph{arXiv preprint arXiv:1910.12744}, 2019.

\bibitem[Saremi(2020)]{saremi2020learning}
Saeed Saremi.
\newblock Learning and inference in imaginary noise models.
\newblock \emph{arXiv preprint arXiv:2005.09047}, 2020.

\bibitem[Saremi and Hyv{{\"a}}rinen(2019)]{saremi2019neural}
Saeed Saremi and Aapo Hyv{{\"a}}rinen.
\newblock Neural empirical {B}ayes.
\newblock \emph{Journal of Machine Learning Research}, 20\penalty0
  (181):\penalty0 1--23, 2019.

\bibitem[Saremi et~al.(2018)Saremi, Mehrjou, Sch{\"o}lkopf, and
  Hyv{\"a}rinen]{saremi2018deep}
Saeed Saremi, Arash Mehrjou, Bernhard Sch{\"o}lkopf, and Aapo Hyv{\"a}rinen.
\newblock Deep energy estimator networks.
\newblock \emph{arXiv preprint arXiv:1805.08306}, 2018.

\bibitem[Swersky et~al.(2011)Swersky, Ranzato, Buchman, Marlin, and
  de~Freitas]{swersky2011autoencoders}
Kevin Swersky, Marc'Aurelio Ranzato, David Buchman, Benjamin~M Marlin, and
  Nando de~Freitas.
\newblock On autoencoders and score matching for energy based models.
\newblock In \emph{ICML}, 2011.

\bibitem[Torralba et~al.(2008)Torralba, Fergus, and Freeman]{torralba200880}
Antonio Torralba, Rob Fergus, and William~T Freeman.
\newblock 80 million tiny images: A large data set for nonparametric object and
  scene recognition.
\newblock \emph{IEEE transactions on pattern analysis and machine
  intelligence}, 30\penalty0 (11):\penalty0 1958--1970, 2008.

\bibitem[V{\'e}rtes and Sahani(2016)]{vertes2016learning}
Eszter V{\'e}rtes and Maneesh Sahani.
\newblock Learning doubly intractable latent variable models via score
  matching.
\newblock Technical report, Gatsby Unit, UCL, 2016.

\bibitem[Vincent(2011)]{vincent2011connection}
Pascal Vincent.
\newblock A connection between score matching and denoising autoencoders.
\newblock \emph{Neural Computation}, 23\penalty0 (7):\penalty0 1661--1674,
  2011.

\end{thebibliography}
\bibliographystyle{alpha}

\end{document}